\documentclass{article}

\usepackage{arxiv}

\usepackage[utf8]{inputenc} % allow utf-8 input
\usepackage[T1]{fontenc}    % use 8-bit T1 fonts
\usepackage{hyperref}       % hyperlinks
\usepackage{url}            % simple URL typesetting
\usepackage{booktabs}       % professional-quality tables
\usepackage{amsfonts}       % blackboard math symbols
\usepackage{nicefrac}       % compact symbols for 1/2, etc.
\usepackage{microtype}      % microtypography
\usepackage{lipsum}
\usepackage{array,multirow}

\usepackage{caption}
\usepackage{threeparttable}
\usepackage [english]{babel}
\usepackage [autostyle, english = american]{csquotes}
\MakeOuterQuote{"}

\title{Theory-based Habit Modeling for Enhancing Behavior Prediction}

\author{
  Chao Zhang\thanks{Correspondence can be sent to: Chao Zhang, Utrecht University, Department of Psychology, PO BOX 80140, 3508 TC, Utrecht, The Netherlands; e-mail: c.zhang@uu.nl or chao.zhang87@gmail.com. The majority of this work was done when the first author was a PhD student at Eindhoven University of Technology. We thank Bo Liu and Hanne Spelt for their contributions to the data collections in Study 1 and Study 2 respectively.  Data, analysis script, and other materials can be found in a Open Science Framework repository: https://osf.io/adkb4/. } \\
  Human-Technology Interaction Group\\
  Eindhoven University of Technology\\
  Eindhoven, the Netherlands \\
  \texttt{chao.zhang87@gmail.com} \\
   \And
 Joaquin Vanschoren \\
  Department of Mathematics and Computer Science\\
  Eindhoven University of Technology\\
  Eindhoven, the Netherlands \\
  \texttt{j.vanschoren@tue.nl } \\
  \And
 Arlette van Wissen \\
  Brain, Behavior, and Cognition\\
  Philips Research\\
  Eindhoven, the Netherlands \\
  \texttt{arlette.van.wissen@philips.com } \\
  \And
 Daniël Lakens \\
  Human-Technology Interaction Group\\
  Eindhoven University of Technology\\
  Eindhoven, the Netherlands \\
  \texttt{d.lakens@tue.nl } \\
  \And
 Boris de Ruyter \\
  Brain, Behavior, and Cognition\\
  Philips Research\\
  Eindhoven, the Netherlands \\
  \texttt{boris.de.ruyter@philips.com} \\
  \And
 Wijnand A. IJsselsteijn \\
  Human-Technology Interaction Group\\
  Eindhoven University of Technology\\
  Eindhoven, the Netherlands \\
  \texttt{w.a.iJsselsteijn@tue.nl} \\
}
\usepackage{graphicx}
\graphicspath{ {./} }

\begin{document}
\maketitle

\begin{abstract}
Psychological theories of habit posit that when a strong habit is formed through behavioral repetition, it can trigger behavior automatically in the same environment. Given the reciprocal relationship between habit and behavior, changing lifestyle behaviors (e.g., toothbrushing) is largely a task of breaking old habits and creating new and healthy ones. Thus, representing users’ habit strengths can be very useful for behavior change support systems (BCSS), for example, to predict behavior or to decide when an intervention reaches its intended effect. However, habit strength is not directly observable and existing self-report measures are taxing for users. In this paper, built on recent computational models of habit formation, we propose a method to enable intelligent systems to compute habit strength based on observable behavior. The hypothesized advantage of using computed habit strength for behavior prediction was tested using data from two intervention studies, where we trained participants to brush their teeth twice a day for three weeks and monitored their behaviors using accelerometers. Through hierarchical cross-validation, we found that for the task of predicting future brushing behavior, computed habit strength clearly outperformed self-reported habit strength (in both studies) and was also superior to models based on past behavior frequency (in the larger second study). Our findings provide initial support for our theory-based approach of modeling user habits and encourages the use of habit computation to deliver personalized and adaptive interventions. 
\end{abstract}

% keywords can be removed

\section{Introduction}
\label{intro}
Behavior change support systems (BCSS) are digital systems that support users to change their behaviors in desirable ways by using various intervention techniques \cite{oinas2013foundation,lathia2013smartphones}, including education, persuasion, reminders, contingent rewards, and self-monitoring \cite{abraham2008taxonomy}. In many application domains where behaviors are repeated frequently, such as promoting healthy lifestyles, one of the challenges for successful change is the task of breaking bad old habits and forming healthy new habits \cite{gardner2019habit,karppinen2018opportunities,pinder2018digital}. Habitual behaviors are characterized as automatic responses triggered by cues in the environment (e.g., eating crisps when watching TV) or by goals activated in one's working memory (e.g., using a bike when commuting to work) \cite{sheeran2005goal,wood2007new}. The lack of deliberations about behavioral consequences makes habitual behaviors to persist even when a contradicting goal or intention is formed \cite{dickinson1985actions}. For example, in a press conference on curbing the spread of COVID-19, right after Mark Rutte, the Prime Minister of the Netherlands, told people not to shake hands, he welcomed a medical expert on stage and shook his hand. On the bright side, when a good habit is formed, it helps behavioral maintenance and prevents relapses. Modeling users' habits can potentially increase the effectiveness of BCSS. 

Although the term "habit" is intuitively understood by most people, it is important to clarify what we mean by "habit" in our paper. In the field of ubiquitous computing, modeling habits usually refers to the modeling of users' actual behaviors, i.e., detecting and recognizing recurrent behavioral patterns and routines \cite{kalantarian2015monitoring,meng2017towards,shoaib2015towards}, sometimes contingent on specific user contexts \cite{banovic2016modeling}. In contrast, based on psychological theories \cite{sheeran2005goal,wood2007new,verplanken2018psychology,marien2019studying,wood2016psychology}, we define habits  as the cognitive associations between user behaviors and the triggering user contexts, thus separating habits from habitual behaviors themselves. Habits are strengthened through context-dependent behavior repetitions and in turn increase the probability that the  behavior is performed in the same context. Our modeling approach harnesses this reciprocal causal relationship for behavior prediction and personalized intervention.  

Modeling the habit strength of a particular user behavior can benefit BCSS in at least two ways. First, assuming a causal effect of habit on behavior, knowing the habit strength can assist a system to predict a user's behavior more accurately. Accurate behavior prediction is the basis for personalizing interventions, for example, sending a reminder when the system predicts that the user is unlikely to perform the desirable behavior on their own. Second, it is widely acknowledged that reminders in many so-called "habit-formation" apps induce behavior repetition but hinder the formation of real habits that are supposed to be connected to environmental cues \cite{renfree2016don,stawarz2014don,stawarz2015beyond}. Thus, representing habit strength as a cognitive state enables a system to distinguish genuine context-driven habitual behaviors from repeated behaviors that are simply prompted by digital systems. It also allows a system to decide when to withdraw proactive interventions on a specific behavior, knowing from the model that the user's behavior will likely be maintained by the strong habit alone. 

In psychology, habit strength is often measured by the Self-report Habit Index (SRHI) \cite{verplanken2003reflections} or its behavioral automaticity sub-scale \cite{gardner2012towards}. Although these methods can be implemented in a BCSS as daily questionnaires, it burdens users a lot and may interfere with primary intervention techniques. Recently developed computational models provide a new approach of quantifying habit strength based on observable behavior and contexts \cite{klein2011computational,miller2019habits,psarra2016bounded,tobias2009changing}, but these models have not been extensively validated in real-world behavior change interventions. In this paper, we evaluated one habit-learning model using two field intervention studies on dental hygiene behavior, testing whether computing habit strength contributes to more accurate behavior prediction, when compared with theory-free predictive models. Enhanced prediction performance also empirically validates the model for representing users' habit strengths. 

In the remainder of the paper, we start with the theoretical background of our work, followed by the overall modeling and evaluation approach. Next, the data-collection method and results of the two field studies are presented. The paper concludes with a general discussion, including implications for designing more personalized BCSS. 

\section{Theoretical Background}
\label{background}
\subsection{The Psychology of Habit}
\label{theory}
Habits can be best understood in the context of human goal-directed behavior. To say a behavior is habitual implies that the behavior is performed merely due to its repetitions and instrumental values in the past (e.g., shaking hands as a social norm), but may be at odds with one's current goals (e.g., to prevent the spread of virus). The most powerful demonstrations of habitual behavior come from animal and human instrumental learning experiments \cite{dickinson1985actions}, in which extensively repeated behavioral responses become insensitive to sudden changes in the immediate decision-making environments. Although a functional dissociation between habitual and goal-directed behavior is well-established \cite{yin2006role}, there remains a strong dispute between a \textit{value-based} and a \textit{value-free} view regarding the specific cognitive mechanisms underlying habit learning and its control over overt behavior.

The value-based account of habit conceptualizes habit learning as a form of value-free reinforcement learning \cite{daw2005uncertainty,keramati2011speed}. Through repeated choices, an organism learns the goal-satisfying values of different behavioral options and uses the "cached" action-values to make new decisions, without resorting to a model of its environment. This contrasts with goal-directed learning or model-based reinforcement learning, which informs decisions based on an up-to-date model of the environment. Theories in the value-based account also hypothesize that after extensive training, a cognitive system based on habit learning controls behavior thanks to its simplicity and efficiency. 

However, the value-based account conflicts with the more traditional value-free view of habit in psychology, which has its root in the classic distinction between "\textit{law of exercise}" and "\textit{law of effect}" by Edward Thorndike \cite{thorndike1932fundamentals}. FOr the value-free account, a habit only relates to the repetition of a behavior (exercise), but not the outcomes of executing the behavior (effect). In modern terms, as a by-product of goal-directed learning, a habit is a learned cognitive association between a behavior and its triggering context or goal, strengthened by repeated behavior executions \cite{wood2007new,marien2019studying,wood2016psychology}. When the same context is encountered or the same goal is activated, this association immediately brings a representation of the behavior into one's working memory \cite{tobias2009changing} or enhances the baseline preference signal of the behavior in decision-making \cite{roe2001multialternative,zhang2019towards}. The present research follows the value-free account of habit learning and makes use of its computational models.

\subsection{Computational Models of Habit Learning}
\label{computational}
Following the value-free view, four computational models have been proposed to account for the relationship between behavior repetition and habit strength \cite{klein2011computational,miller2019habits,psarra2016bounded,tobias2009changing}. As these models were developed by researchers from very different fields, they haven't been reviewed in the same context. While a detailed comparison between the models is beyond the scope of the paper, it suffices to say that they all followed the value-free account and were inspired by the Hebbian learning principle in neuroscience \cite{hebb1949organization}. In a network of cognitive nodes representing behaviors and contextual cues, the links between pairs of behaviors and cues are strengthened when the two nodes are activated at the same time (i.e., a behavior is performed in that particular context). 

Figure 1 shows the mathematical equations of these models and a simulation of how habit strength changes over time in a prototypical scenario with plausible parameter values of each model. Despite their differences, all models produce a similar pattern for the dynamics of habits: habit strength increases over time when the behavior is performed consistently but the rate of growth decreases so that habit strength approaches a plateau. When the behavior is not performed, habit strength decays proportionally \footnote{Tobias's equation does not simulate habit decay while in isolation, but one should note that the equation for habit formation is meant to be used together with other equations in a complex model of behavior change.}. These basic patterns are consistent with the empirical data of habit formation in a field study where participants reported their habit strengths using the SRHI \cite{lally2010habits}.

\begin{figure}[ht]
  \centering
  \includegraphics[width=0.8\linewidth]{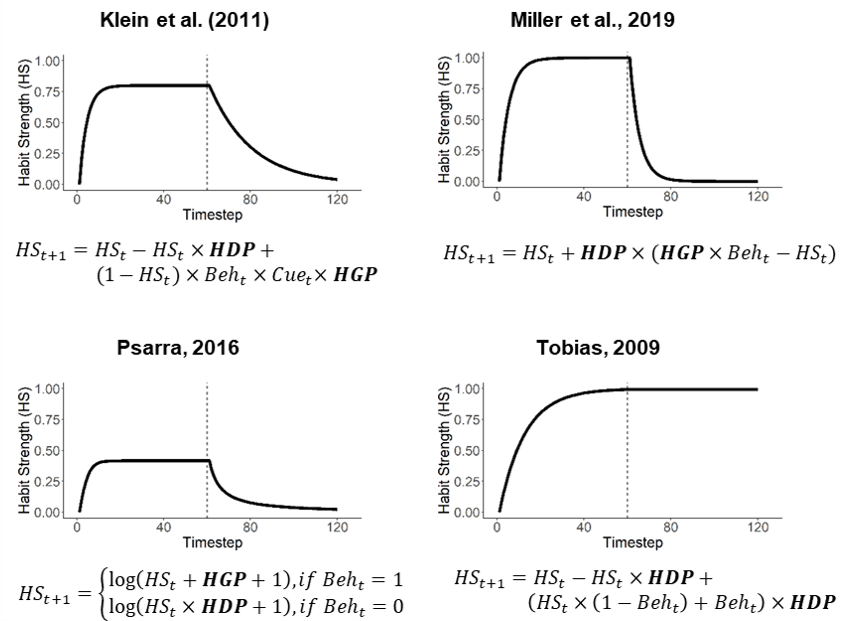}
  \setlength{\belowcaptionskip}{-5pt}
  \caption{Equations of different computational models of habit learning and their simulation results under a simple scenario. (HS = habt strength, Beh = behavior, HDP = habit decay parameter, HGP = habit gain parameter. Note that original parameter names were unified by the authors for clarity of presentation, but their exact meanings are bounded by each of the equations.) }
  \label{fig:model}
\end{figure}

\subsection{Habit Formation and Memory Process}
\label{memory}
The automaticity of habits in daily environments has been partly attributed to their facilitation role in memory processes \cite{psarra2016bounded,tobias2009changing}. Unlike laboratory studies on habits (e.g., \cite{dickinson1985actions}) where behavioral options are presented to the participants, in daily lives people have to recall certain behavioral options before they can choose among them \cite{kamphorst2015option}. When a habit is strong, the learned association between a behavior and a context ensures that when the same context is encountered, the behavior is activated in one's working memory as a choice option \cite{wood2007new}. According to \cite{psarra2016bounded}, habitual behaviors are hard to control because habitual options are recalled and evaluated first. When habitual options are sufficiently satisfying, people may act on them immediately without trying to recall more options.

When a habit is still weak, the newly learned behavior can often be "forgotten" in the moments of generating behavioral options. Thus, to ensure early behavioral repetitions, memory aids such as digital reminders are often needed. In addition to modeling habit formation, other researchers also proposed a computational model of how memory accessibility of behavioral options changes over time \cite{tobias2009changing}. Like any other memory processes, the accessibility of a behavioral option decays gradually over time but can be restored upon receiving reminders or when the behavior is executed. Other unobservable factors, such as the mental rehearsal of an option \cite{einstein1996retrieval}, also influence accessibility but their effects are integrated into a decay parameter in \cite{tobias2009changing}. We adopted Tobias's equation, which is defined formally in the next section. 

\section{Modeling and Evaluation Approach}
\label{approach}

\subsection{Computing Habit Strength and Memory Accessibility}
\label{computing}
Based on the theories and computational models reviewed, we focus on two cognitive quantities that can be computed by a digital system. Of our primary interest, the habit strength of a target behavior for a user in a behavior change process is computed. In principle, any of the 4 computational models reviewed above \cite{klein2011computational,miller2019habits,psarra2016bounded,tobias2009changing} can be used for this computation, but we chose to use the equation in \cite{klein2011computational} since it is the only model that can match the empirical observation that habit growth is faster than habit decay \cite{lally2010habits}. The equation with a habit decay parameter (\textit{HDP}) and a habit gain parameter (\textit{HGP}) is as the following:

\begin{equation} HS_{t+1} = HS_{t} - HS_{t} \times \textit{HDP} + (1 - HS_{t}) \times Beh_{t} \times Cue_{t} \times \textit{HGP} \end{equation}

The equation implies that given an initial habit strength of a user (\textit{$HS_0$}), the subsequent habit strength at any time point (\textit{$HS_t$}) can be computed as long as the past occurrences of behavior (\textit{Beh}) and cues (\textit{Cue}) are known. In an empirical study or a behavior change application, users can be asked to self-report their habit strengths at the beginning and the self-reported values (scaled to [0, 1]) can be used as initial values. Both actual behavior and environmental cues can be potentially monitored by sensors in a BCSS. In the current research, we make a simplifying assumption that users always perform the target behavior in the same context (i.e., participants in our studies always brushed teeth in their own bathrooms and at similar time), so the variable {$Cue_t$} is always 1. 

In addition to habit strength, the memory accessibility of a behavioral option can be computed using the equation in \cite{tobias2009changing}. Accessibility (\textit{Acc}) decays naturally as a natural memory process, but can be enhanced by behavior executions (\textit{Beh}) and external reminders (\textit{Rem}). The equation controlled by three free parameters – accessibility decay parameter (\textit{ADP}), accessibility gain parameter with behavior execution (\textit{$AGP_{beh}$}), and accessibility gain parameter with reminder (\textit{$AGP_{rem}$}), is as the following:

\begin{equation} Acc_{t+1} = Acc_{t} - Acc_{t} \times \textit{ADP} + (1 - Acc_{t}) \times (Beh_{t} \times \textit{$AGP_{beh}$} + Rem_{t} \times \textit{$AGP_{rem}$}) \end{equation}

When a user is persuaded by a BCSS to learn a new behavior, the initial value of memory accessibility (\textit{$Acc_0$}) of the target behavior can be assumed to be 1 (maximum). Subsequent memory accessibility can be easily updated by monitoring actual behavior and reminders sent by the digital system itself. For simplification, any procedure used in our empirical studies (e.g., face-to-face meeting, email communication, etc.) that reminded participants of the target behavior was assumed to restore memory accessibility by the same amount controlled by a single parameter \textit{$AGP_{rem}$}. 

\subsection{Using Computed Variables in Predictive Modeling}
\label{predictive}
The primary goal of the current research is to evaluate the utility of computing habit strength and memory accessibility in a use case of behavior prediction. In a behavior change intervention, predicting future behavior based on information already collected is often an important and challenging task. For example, when a user is prompted by a BCSS to brush teeth every morning, it is a meaningful task to predict whether the user will brush their teeth the next morning (also known as a 1-step forecast) based on all the system knows about the user at that point. A conventional approach for behavior prediction in psychology relies on self-reported behavioral determinants measured by periodical surveys (survey model, see Figure 2a), such as attitude, intention, and self-report habit strength \cite{gardner2015review}. Another method is simply to use past behavior to predict future behavior, for example, by counting how many times the user brushed teeth in the morning prior to the to-be-predicted date (past-behavior model, see Figure 2b). Instead of these two approaches, the system can also compute habit strength and memory accessibility based on historical data (past behavior, reminder, etc.) and use the computed theoretical quantities to predict future behavior (theory-based model, see Figure 2c). Computing the theoretical quantities is useful if the \textit{theory-based model} predicts future behavior more accurately than the \textit{past-behavior model} and at least as accurate as the \textit{survey model}, given that it bypasses the need to burden users with questions. Note that we focus on comparing the relative performance of the models rather than optimizing absolute performance.

\begin{figure}[ht]
  \centering
  \includegraphics[width=0.7\linewidth]{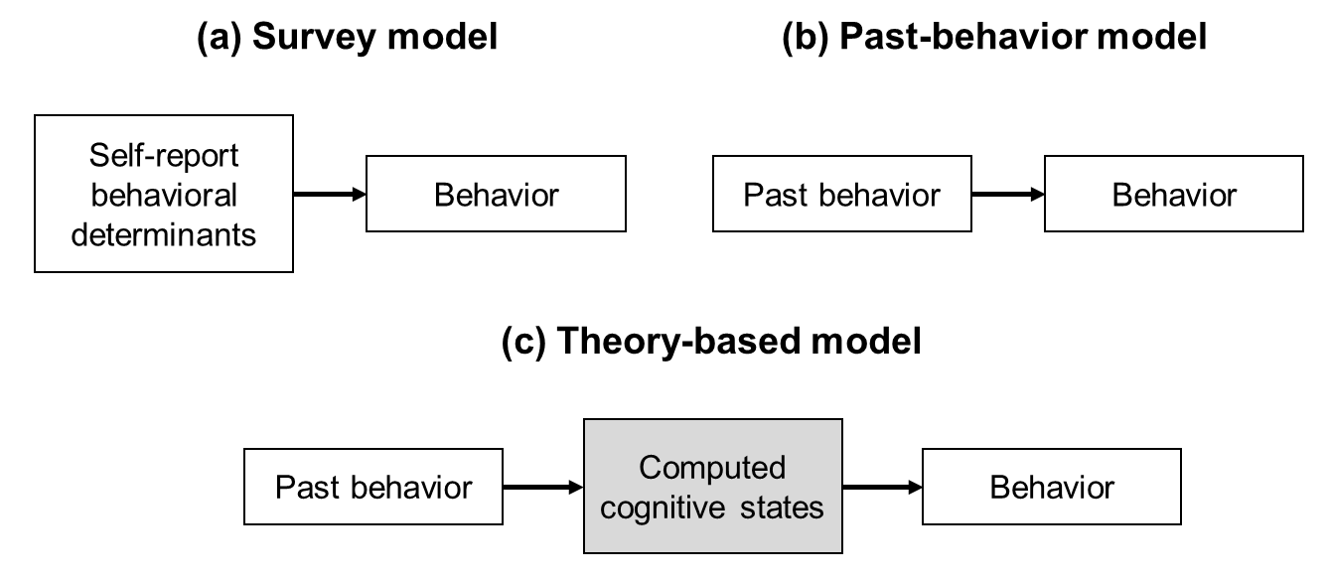}
  \setlength{\belowcaptionskip}{-5pt}
  \caption{Three different modeling approaches: (a) survey model; (b) past-behavior model; (3) theory-based model.}
  \label{fig:approach}
\end{figure}

To fulfill the research goal, we conducted two intervention studies on dental health behavior where participants were trained to brush their teeth twice a day for about three weeks. Participants' brushing behavior was monitored by sensors and their attitude towards toothbrushing and self-report habit strength were measured once a week. We chose to study toothbrushing behavior because of its relative simplicity, context stability (e.g., usually in the bathroom at home), and high occurrence frequency. 

\section{Method}
\label{data}

\subsection{Participants}
\label{participant}
\paragraph{Study 1} Forty healthy university students or young workers were recruited through a local participant database and personal network. The main inclusion criterion was that they used to only brush their teeth once a day (or at least rarely brushing twice), and the criterion was checked by personal communication with the participants. The sampling consisted of 26 males and 14 females, and the average age was 24.48 (\textit{SD} = 3.13, median = 24). Eight participants were randomly selected and awarded 25 euros. The study was reviewed and approved by an ethical review board at Eindhoven University of Technology.

\paragraph{Study 2} Study 2 was conducted in collaboration with Philips Research. Seventy-nine adults were recruited through a recruitment agency contracted by Philips. A lenient main criterion was used that the participants used to brush only once a day, or they usually brushed less than two minutes for each session. Other criteria include that they were between 18 and 60 years old, understood Dutch, and were manual toothbrush users. The eventual sample consisted of 41 females and 37 males (1 chose “other”), with ages between 20 and 63 years old (mean = 39.63, median = 38, \textit{SD} = 10.97). Most participants were healthy, except that one suffered from cystic fibrosis and one from narcolepsy. Participants were paid 80 euros by the recruitment agency. The study was reviewed and approved by the Internal Committee on Biomedical Experiments (ICBE) at Philips Research.

\subsection{Design and Procedure}
\label{design}
\paragraph{Study 1} Participants were enrolled in a 4-week intervention program during which they were persuaded to change their oral health routine from brushing teeth once a day to brushing twice a day. The main outcome variable was whether they complied with the new target brushing behavior (i.e., brushing also in the morning or in the evening) on each day during the study period. At the beginning, a face-to-face meeting was held between the experimenter and each participant. During this meeting, participants were introduced to the study and the intervention, signed a consent form, and were given a sensor to be attached to their own toothbrush. After participants returned home, their toothbrushing behaviors were monitored by the sensors for 3 weeks, and at the end of the third week they returned the sensor to the experimenter. Reminders for the target brushing behaviors were sent daily in the first week using a self-programmed mobile app, every other day in the second week, and were dismissed in the third and fourth week. at the end of each week, a short survey was sent using the same app to ask questions about attitude and habit strength.

\paragraph{Study 2} Participants were enrolled in a multi-phase intervention program during which they were persuaded to  develop an optimal oral health routine of two brushing sessions that last for at least 2 minutes (or at least a 4-minute brushing daily). The main outcome variable was whether they brushed their teeth twice a day or not. At the beginning, participants came to the lab in groups of 10-15 for an introduction session, in which general study information and procedure were explained, but not the specific intervention. Also in the meeting, participants were offered new manual toothbrushes with sensors attached, and were asked to sign a consent form and to complete the first survey. After the baseline period of about 5-10 days, they were invited back to the lab for the intervention session individually. They were shown presentations about oral healthcare, and were exposed to the intervention target of brushing twice a day for at least 4 minutes. During the lab session, physiological data from the participants were recorded for purposes unrelated to this paper (see \cite{spelt2020persuasion}). The second and third survey, with mostly identical questions, was completed by the participants before and after the lab session. After the lab session, participants returned home and were monitored for a follow-up period that led to a total of approximately 3 weeks. Two additional surveys were sent by e-mail in the middle and at the end of the follow-up period. 

\subsection{Measurements}
\label{measurement}
\paragraph{Toothbrushing behavior} Participants' toothbrushing behavior was measured by the Axivity AX3 sensors attached to the lower-end of their toothbrush grips (see Figure 3). The Axivity AX3 sensor is a 3-axis accelerometer developed by Newcastle University specifically for scientific research on human movements \cite{doherty2017large}. Constrained by the memory space of the device, the sampling frequency was set at 50 Hz to ensure the storage of data for three weeks. The sensitivity range for accelerations was set at $\pm$8g. The sensor was waterproof, and a fully-charged sensor could work for 3 weeks without additional charges. Participants in both studies also self-reported on how many days of the previous week they brushed their teeth in the morning/evening (Study 1) or brushed teeth twice a day for at least 2 minutes each time (Study 2).

\begin{figure}[ht]
  \centering
  \includegraphics[width=0.45\linewidth]{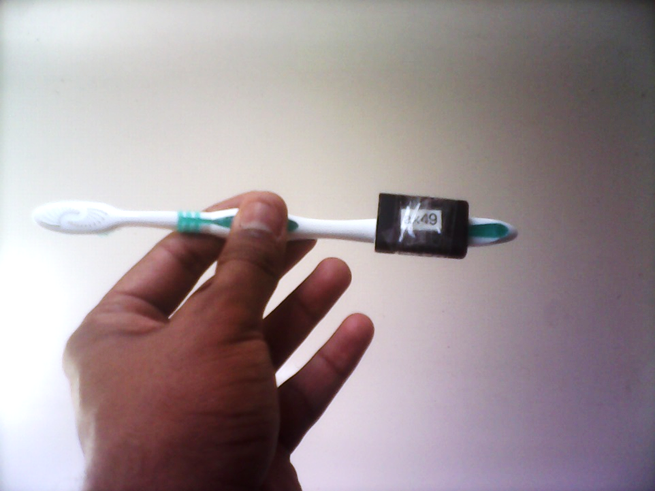}
  \setlength{\belowcaptionskip}{-5pt}
  \caption{An example of how the Axivity AX3 accelerometer was attached to a toothbrush in the studies.}
  \label{fig:sensor}
\end{figure}

\paragraph{Habit strength} Habit strength was measured using the 4-item Self-report Behavior Automaticity Index (SRBAI) with 7-point response scales \cite{gardner2012towards}. It assessed behavioral automaticity by prompting participants to rate their agreements with descriptions of performing a target behavior (e.g., \textit{“Behavior X is something…”}), including \textit{“I do automatically”}, \textit{“I do without having to consciously remember”}, \textit{“I do without thinking”}, and \textit{“I start doing before I realize I am doing it”}. The target behavior in Study 1 was \textit{“brushing teeth in the morning”} or \textit{“brushing teeth in the evening”}, depending on which behavior was not performed by each participant before the study. In Study 2, because of the lenient inclusion criterion, the behavior was more generally phrased as \textit{“brushing teeth twice a day and in total at least 4 minutes”}. Internal reliabilities of the SRBAI were very high in both Study 1 (Cronbach’s $\alpha$ = 0.95) and Study 2 (Cronbach’s $\alpha$ = 0.94). These items were translated into Dutch in Study 2.

\paragraph{Attitude} Attitude was measured using 7-point semantic differential scales that were typically used in studies that followed the Theory of Planned Behavior \cite{verplanken1997habit}. Four items were used in Study 1 (bad – good, useless – useful, harmful – beneficial, unpleasant – pleasant), while in Study 2 three more items were added (foolish – wise, unhealthy – healthy, difficult – easy). We also made a common distinction between instrumental attitude and affective attitude \cite{tobias2009changing}, because inter-item correlations and factor analysis clearly suggested that there were two separate factors. Instrumental attitude focuses on how a behavior satisfied instrumental goals, such as health benefits in the context of dental behaviors, while affective attitude taps more onto the emotional aspects of the experience relating to the behavior (e.g., comfort of brushing, effort spent on brushing). The affective attitude score was based on a single item in Study 1 (unpleasant – pleasant) and the average score of two items in Study 2 (unpleasant – pleasant, difficult – easy). Internal reliabilities (Cronbach’s $\alpha$) for instrumental attitude were 0.94 and 0.93 for the two studies, while affective attitude also had a satisfying internal reliability of 0.71 in Study 2. The attitude items were translated into Dutch in Study 2.

\subsection{Pre-processing}
Pre-processing was performed to transform the raw 3-axis accelerometer data to behavioral data at the day-level (i.e., brushing twice or not on a specific day). The same procedure was used in both studies, which included the following steps: converting 3-axis signals to signal vector magnitudes (SVM), extracting brushing episodes, and classifying episodes to determine the main outcome variables.
\label{preprocessing}

\paragraph{Converting 3-axis signal to SVM} The first step was to compute SVM based on the raw three-axis accelerometer data, according to the equation below: SVM provided a summarized movement magnitude measure by combining the acceleration information from the x, y, and z axis, and down-sampling the 50 Hz raw data to magnitude measured at 1 Hz (n = 50 in the equation above). Figure 4a shows one participant’s data after SVM-transformation, where each data point (dot) represents the average movement magnitude in each 1-second time window. This processing was done using a built-in SVM algorithm Open Movement v1.0.030, the default software for the Axivity AX3 sensor.

\paragraph{Extracting brushing episodes} From Figure 4a, it was clear that brushing episodes could even be visually identified (the spikes) when the data were clean, but not when there was noise caused by other movements. Given this problem, a threshold-based algorithm was first used to scan the data sequentially to efficiently extract all potential brushing episodes (see the classified data points in Figure 4b), and then a manual check was performed to exclude “invalid” episodes. The details of this step can be found in \cite{zhang2019towards}. 

\begin{figure}[ht]
  \centering
  \includegraphics[width=0.85\linewidth]{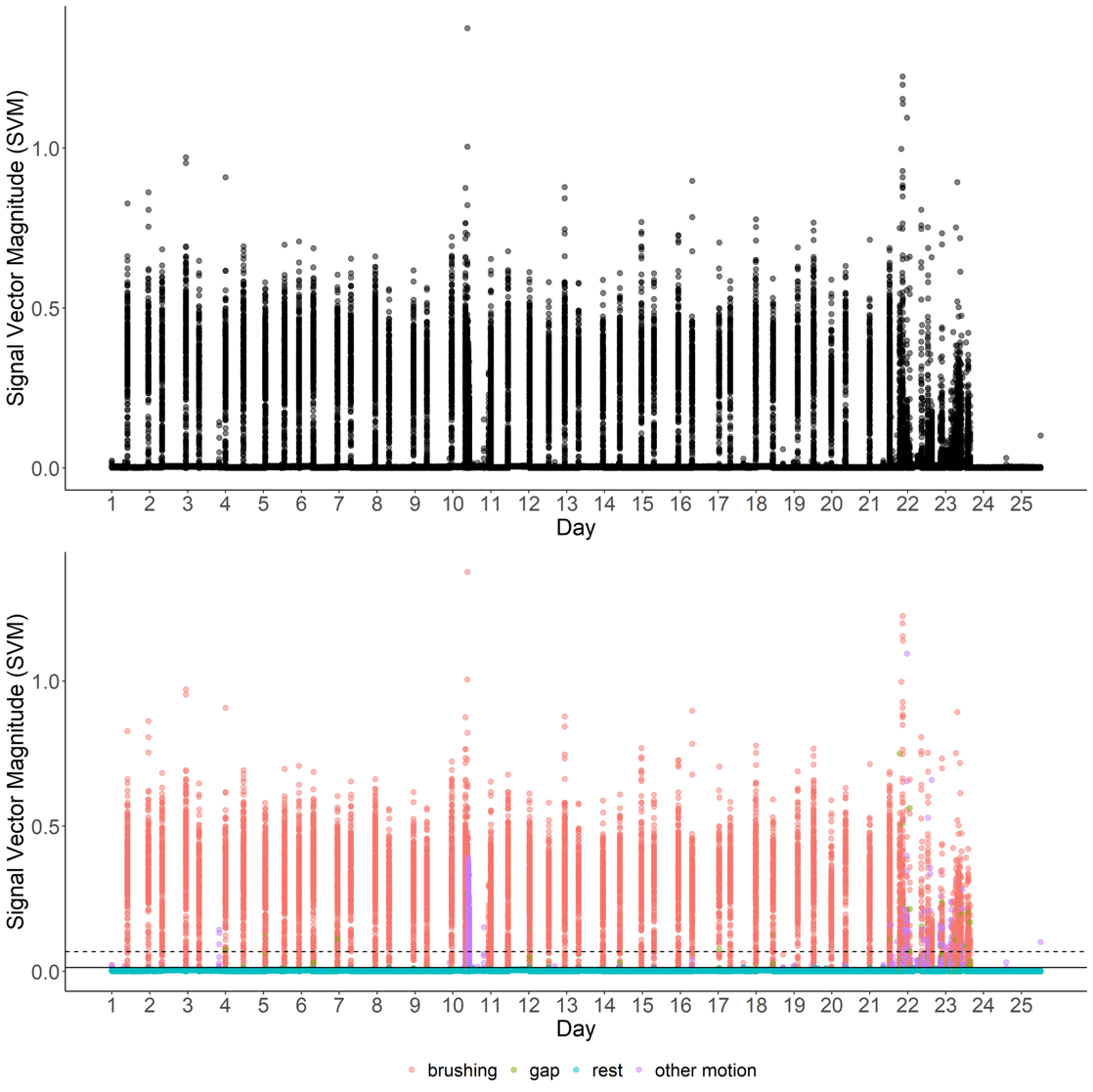}
  \setlength{\belowcaptionskip}{-5pt}
  \caption{(a) Sensor data in signal vector magnitude; (b) Sensor data classified after applying the threshold-based algorithm.}
  \label{fig:data}
\end{figure}

\paragraph{Classifying episodes to create the main outcome variables} The remaining episodes were then classified into 6 categories based on the starting time of the episodes: \textit{morning} (5:00 – 12:00), \textit{morning-afternoon} (12:00 – 15:00), \textit{afternoon} (15:00 – 19:00), \textit{afternoon-evening} (19:00 – 21:00), \textit{evening} (21:00 – 24:00), and \textit{overnight} (0:00 – 5:00). The final episode-level data may contain more than one episode for each time category on each date. At the data level, two variables – \textit{morning brushing} and \textit{evening brushing} – were created, and their values (0 or 1) were determined by searching in the relevant categories on the same date to see if any episode existed. For \textit{morning brushing}, category \textit{morning} was searched first, and if no episode was found, category \textit{morning-afternoon} was searched. For \textit{evening brushing}, categories evening and overnight were searched first, and if no episode was found, category \textit{afternoon-evening} was searched. When there were known or unknown events that caused noise in the data in a certain period, the values for the two brushing variables were coded as missing data. Eventually, at the day level, dichotomous indicators (0 or 1) for \textit{the target brushing behavior} and for \textit{brushing twice} were used as the outcome variable in Study 1 and Study 2 respectively.

\subsection{Model Comparison}
The target for prediction was the brushing behavior on the next day, with the occurrence of brushing as the \textit{negative cases} and the absence of brushing as the \textit{positive cases}. They were coded in this way because for real applications a potentially more important goal would be to detect the positive cases, i.e., the days on which the brushing behavior was likely to be omitted. The theory-based computational approach would be considered valuable if it led to models that performed better than models based simply on past behavior or on weekly self-reported variables. Specifically, models with 4 different feature sets were compared:

\begin{itemize}
  \item \textit{Survey model}: The primary features in the survey model were the variables measured by weekly surveys, including \textit{instrumental attitude}, \textit{affective attitude}, and \textit{self-reported behavioral automaticity}. In addition, the \textit{occurrence of lab sessions} (including the introduction meeting in Study 1) and the \textit{occurrence of reminders} (including notifications and e-mails for surveys) were also included as features.
  \item \textit{Past-behavior model}: The primary feature in this model was the past behavior rate until the day of the last observation. For example, if the brushing behavior on the $11^{th}$ day was to be predicted, the brushing rate in the last 10 days (e.g., 0.8) would be the value for this variable. For the first day, past behavior rate was set to 0 in Study 1, as participants self-reported to rarely brush in the morning or in the evening. In Study 2, the self-reported behavior rates in the previous week were used for the initial values. Again, the \textit{occurrence of lab sessions} and the \textit{occurrence of reminders} were also included as features.
  \item \textit{Theory-based model}: This was the model of our interest that includes only computed \textit{habit strength} and \textit{accessibility} as features. 
  \item \textit{Combined model}: The combined model included features in both the past-behavior model and the theory-based model. It was used mainly to evaluate whether combining past behavior and computed cognitive features could further boost prediction performance. 
\end{itemize}

For each model type, three common statistical learning algorithms were used, namely logistic regression, support vector machine, and random forest. In total, this resulted in 12 models (4 model types $\times$ 3 algorithms) to be trained and tested.

Two different approaches were used to compare model performance. First, a two-level hierarchical k-fold cross-validation procedure was used on each of the two data sets separately (see Figure 5).  For each data set, all observations were divided into \textit{k} non-overlapping groups (with the restriction that one participant’s data were always in only one group), so that 1 group was reserved for model testing, and the remaining \textit{k}-1 groups were used for training in each round (the outer loop). Because tuning was needed for both the free parameters in the equations of HS and Acc and the hyperparameters for support vector machine and random forest, the training set in each round was further divided, with 1 group reserved as the test set for parameter tuning and the remaining \textit{k}-2 groups as the training set for parameter tuning (the inner loop). For each free parameter in the theory-based equations, a 1000-step random search was used, and in each step a random value was drawn from a uniform distribution between 0 and 1. For the hyperparameters, grid-search was used to swipe the parameter space as defined in Table 1. These parameter values were optimized to obtain the best overall prediction performance in the inner cross-validation loop, indicated by area under curve (AUC) in receiver operating characteristic (ROC) curves. Due to the sample size difference between the two studies, 9 folds were used for Study 1 (4 participants in each group) and 5 folds were used for Study 2 (15 participants in each group), in order to have sufficient data for training. 

\begin{figure}[ht]
  \centering
  \includegraphics[width=0.95\linewidth]{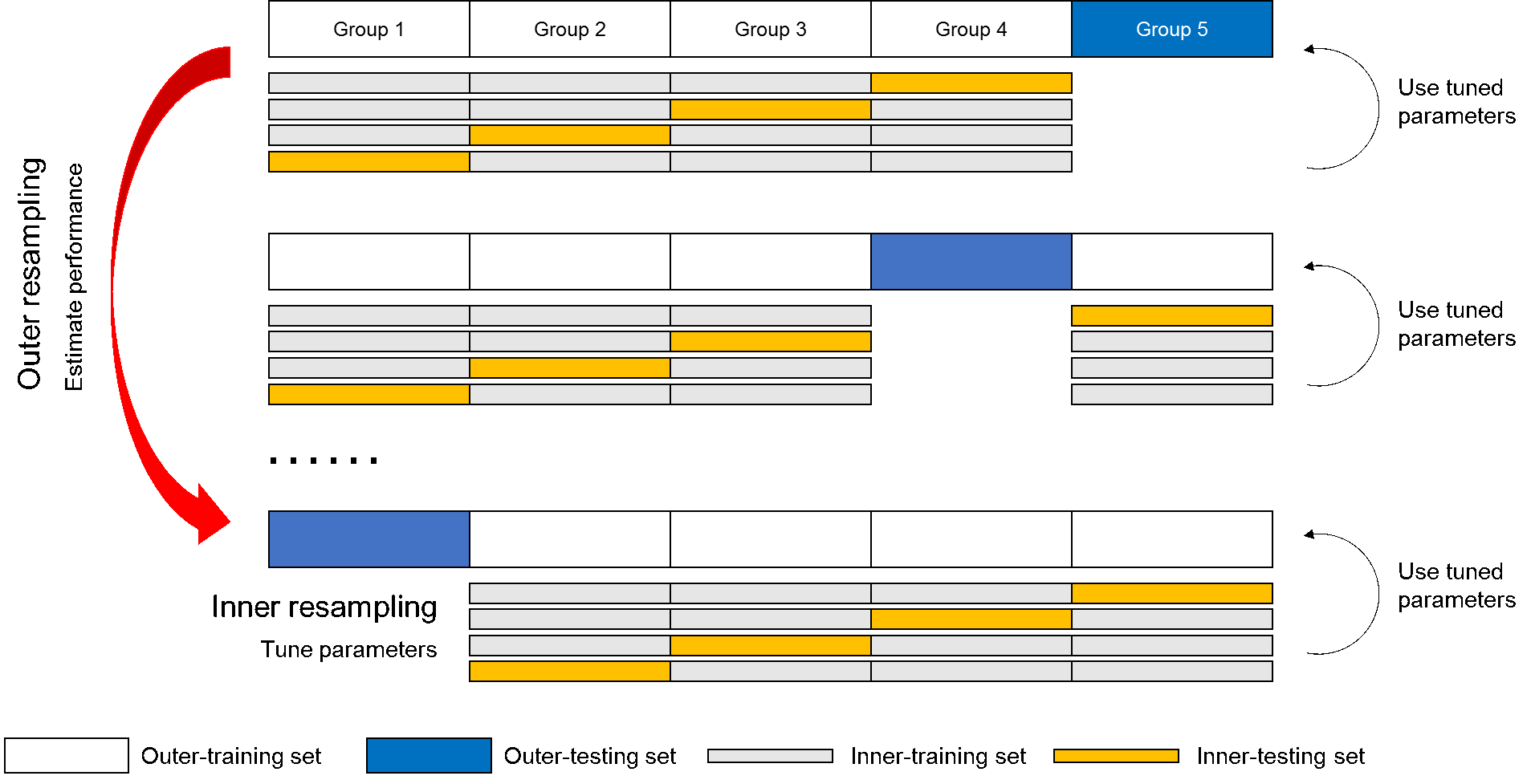}
  \setlength{\belowcaptionskip}{-5pt}
  \caption{An illustration of the nested cross-validation procedure used (it shows the 5-fold scenario for Study 2, but the same idea applies to Study 1).}
  \label{fig:cv}
\end{figure}

\begin{table}
\centering
\caption{Range of values used in hyperparameter tuning.}
\label{tab:parameters}
\begin{tabular}{ l l } 
 \toprule
 \textbf{Support vector machine} &  \\ 
 C (regularization parameter) & 0.1, 1, 10, 100, 1000 \\
 $\gamma$ (width parameter of the Gaussian kernel) & 0.0001, 0.001, 0.01, 0.1, 1, 10 \\
 \midrule
 \textbf{Support vector machine} & \\
 \textit{ntree} (number of trees) & 500 \\
 \textit{nodesize} (minimum observations in the terminal nodes) & 1, 4, 16, 64, 256 \\
 \textit{mtry} (number of features used for node split) & 1, 2, ..., $n_{feature}$ \\
 \bottomrule
\end{tabular}
\end{table}

Since we had two similar data sets, in a second approach, we evaluated the ability of each model type to predict new data. This approach was used to evaluate the generalizability of the models, in particular the generalizability of the parameters used to the compute theory-based features (e.g., HGP, ADP). Specifically, one of the two data sets was used to train the models, and the resultant models were used to predict the observations in the other data set. When parameter tuning was required, a k-fold cross-validation was used on the whole training data set, with the same search methods indicated above. Again, 9-fold or 5-fold cross-validation was used when Study 1 or Study 2 was used as the training data set respectively. 

For model comparison, we primarily focused on AUC. Compared with other performance metrics, AUC takes both positive and negative cases into account and is generally considered the best for both balanced and unbalanced data sets \cite{halimu2019empirical}. AUC was also chosen because we were more interested in predicted probabilities of brushing rather than the classifications under a particular threshold. For comprehensiveness, we also report other performance measures computed using the optimal threshold for each model, namely Matthew Correlation Coefficient (MCC), overall accuracy, F-score, true positive rate, false positive rate, precision, and negative prediction value. All analyses were performed in R statistical programming environment (version 3.3.3), with the help of the \textit{mlr} (machine-learning R, version 2.1.3) package \cite{bischl2016mlr}. 

\section{Results}
\label{results}

\subsection{Performance within individual datasets}
\label{cv}

\paragraph{Study 1} Study 1 included 711 non-missing observations for the prediction task, with 376 positive cases (non-brushing) and 335 negative cases (brushing). Thus, the prediction accuracy would be 53\% if a no-skill model always predicts positive cases. Figure 6 shows the testing ROC curves of different models, and Table 2 compares additional testing performance measures of the models (aggregated over cross-validation iterations) \footnote{Thresholds used for the logistic regression models were 0.49 (Survey), 0.5 (PB), 0.56 (Theory), and 0.56 (Combined). Threshold used for the SVM models were 0.46 (survey), 0.56 (PB), 0.55 (Theory), and 0.56 (Combined). Threshold used for the random forest models were 0.52 (Survey), 0.84 (PB), 0.4 (Theory), and 0.68 (Combined).}. All models were able to perform substantially better than the no-skill model, with average accuracy ranging between 64\% and 71\%. Various performance measures indicated that the theory-based models were better than the survey models, but were slightly worse than the past-behavior models. It was also clear that combining the features of the theory-based and past-behavior models did not improve performance any further. In terms of learning algorithms, their results were largely the same, although random forest showed more decline of performance from training to testing set, suggesting some overfitting during training.

\begin{table}
\centering
\caption{Comparison of model performances in predicting testing data (Study 1).}
\label{tab:results1}
\renewcommand{\arraystretch}{1.2}
\begin{tabular}{ c l c c c c c c c c } 
 \toprule
  & & AUC & MCC & Acc & TPR & FPR & Precision & F1-score & NPV \\
 \midrule
 \multirow{4}{*}{\rotatebox[origin=c]{90}{\parbox[c]{1.2cm}{\centering Logistic regression}}} & Survey & 0.660 & 0.274 & 0.639 & 0.668 & 0.394 & 0.655 & 0.661 & 0.619 \\ & PB & 0.758 & 0.432 & 0.706 & 0.604 & 0.179 & 0.791 & 0.685 & 0.649 \\ & Theory & 0.737 & 0.390 & 0.688 & 0.604 & 0.218 & 0.757 & 0.672 & 0.637 \\ & Combined & 0.750 & 0.421 & 0.705 & 0.633 & 0.215 & 0.768 & 0.694 & 0.656 \\ 
 \midrule
 \multirow{4}{*}{\rotatebox[origin=c]{90}{\parbox[c]{1cm}{\centering SVM}}} & Survey & 0.651 & 0.309 & 0.657 & 0.729 & 0.424 & 0.659 & 0.692 & 0.654 \\ & PB & 0.759 & 0.423 & 0.707 & 0.649 & 0.227 & 0.763 & 0.701 & 0.662 \\ & Theory & 0.732 & 0.404 & 0.692 & 0.588 & 0.191 & 0.775 & 0.669 & 0.636 \\ & Combined & 0.741 & 0.416 & 0.702 & 0.625 & 0.212 & 0.768 & 0.689 & 0.652 \\
 \midrule
 \multirow{4}{*}{\rotatebox[origin=c]{90}{\parbox[c]{1cm}{\centering Random forest}}} & Survey & 0.663 & 0.341 & 0.671 & 0.670 & 0.328 & 0.696 & 0.683 & 0.645 \\ & PB & 0.699 & 0.390 & 0.681 & 0.551 & 0.173 & 0.781 & 0.646 & 0.621 \\ & Theory & 0.704 & 0.308 & 0.654 & 0.652 & 0.343 & 0.681 & 0.666 & 0.627 \\ & Combined & 0.708 & 0.356 & 0.662 & 0.521 & 0.179 & 0.766 & 0.620 & 0.604 \\
 \bottomrule
\end{tabular}
\begin{tablenotes}
    \small
    \item Note: Survey = survey model; PB = past-behavior model; Theory = theory-based model; SVM = sup-port vector machine; Acc = accuracy; TPR = true positive rate; FPR = false positive rate; NPV = negative prediction value; MMC = Matthews correlation coefficient.
    \end{tablenotes}
\end{table}

\begin{figure}[ht]
  \centering
  \includegraphics[width=0.95\linewidth]{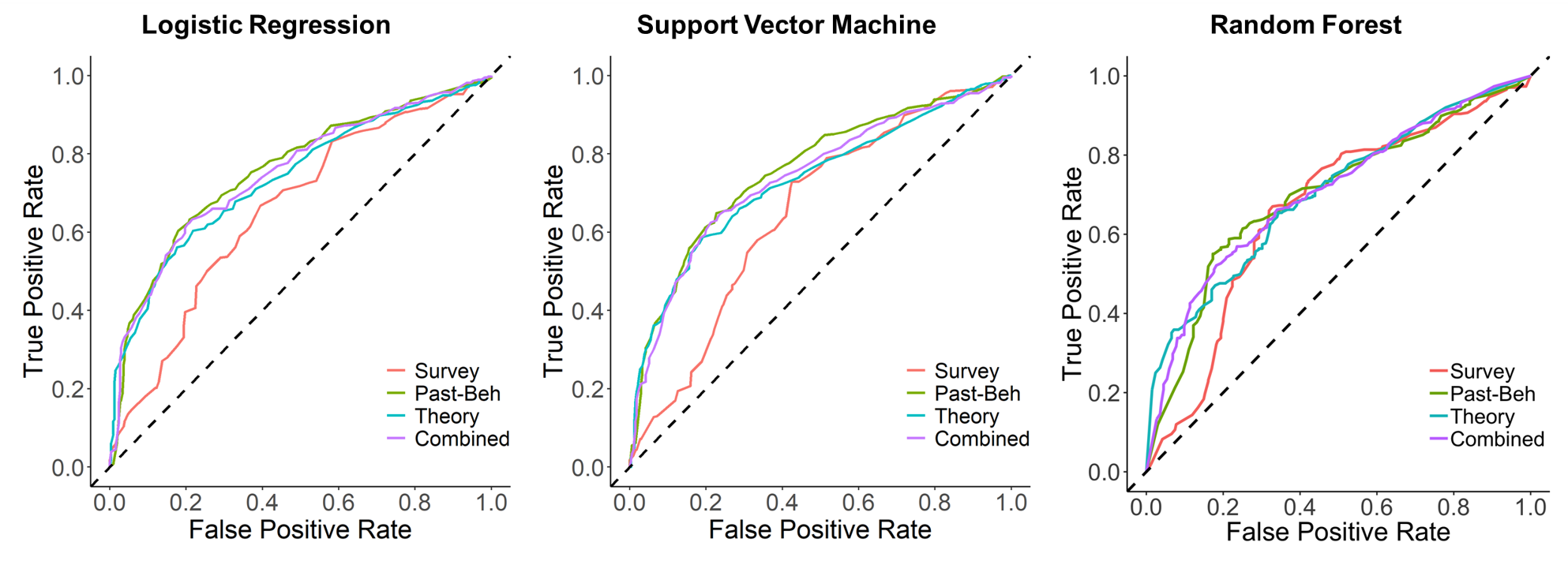}
  \setlength{\belowcaptionskip}{-5pt}
  \caption{Model comparison results of Study 1 based on ROC curves for different models and algorithms.}
  \label{fig:result1}
\end{figure}

\paragraph{Study 2} Study 2 included 1508 non-missing observations for the prediction task, with 557 positive cases (non-brushing) and 951 negative cases (brushing). Thus, the data were less balanced and the prediction accuracy would 63\% if a no-skill model always predicts negative cases. Figure 7 shows the testing ROC curves of different models, and Table 3 compares additional testing performance measures of the models in Study 2 \footnote{Thresholds used for the logistic regression models were 0.43 (Survey), 0.35 (PB), 0.37 (Theory), and 0.38 (Combined). Threshold used for the SVM models were 0.37 (survey), 0.3 (PB), 0.23 (Theory), and 0.32 (Combined). Threshold used for the random forest models were 0.13 (Survey), 0.37 (PB), 0.55 (Theory), and 0.31 (Combined).}. Since the data were more unbalanced (more negative cases due to a higher brushing rate) compared with Study 1, all models were able to predict more accurately, with average accuracy between 64\% and 78\%. In contrast with Study 1, the theory-based models performed much better than the survey models, and also slightly better than the past-behavior models. The models with combined features was arguably the best, although the improvements over the theory-based models were very small. Again, differences between the three algorithms used were very small, but again logistic regression was the best overall.

\begin{table}
\centering
\caption{Comparison of model performances in predicting testing data (Study 2).}
\label{tab:results2}
\renewcommand{\arraystretch}{1.2}
\begin{tabular}{ c l c c c c c c c c } 
 \toprule
  & & AUC & MCC & Acc & TPR & FPR & Precision & F1-score & NPV \\
 \midrule
 \multirow{4}{*}{\rotatebox[origin=c]{90}{\parbox[c]{1.2cm}{\centering Logistic regression}}} & Survey & 0.676 & 0.285 & 0.674 & 0.504 & 0.227 & 0.565 & 0.533 & 0.727 \\ & PB & 0.792 & 0.453 & 0.744 & 0.664 & 0.209 & 0.650 & 0.657 & 0.801 \\ & Theory & 0.815 & 0.491 & 0.767 & 0.641 & 0.160 & 0.701 & 0.670 & 0.800 \\ & Combined & 0.816 & 0.498 & 0.770 & 0.646 & 0.158 & 0.706 & 0.675 & 0.803 \\ 
 \midrule
 \multirow{4}{*}{\rotatebox[origin=c]{90}{\parbox[c]{1cm}{\centering SVM}}} & Survey & 0.668 & 0.264 & 0.642 & 0.612 & 0.341 & 0.513 & 0.558 & 0.744 \\ & PB & 0.794 & 0.443 & 0.723 & 0.749 & 0.291 & 0.601 & 0.667 & 0.828 \\ & Theory & 0.805 & 0.443 & 0.720 & 0.763 & 0.305 & 0.594 & 0.668 & 0.834 \\ & Combined & 0.815 & 0.477 & 0.749 & 0.722 & 0.234 & 0.643 & 0.680 & 0.824 \\
 \midrule
 \multirow{4}{*}{\rotatebox[origin=c]{90}{\parbox[c]{1cm}{\centering Random forest}}} & Survey & 0.661 & 0.281 & 0.651 & 0.618 & 0.329 & 0.524 & 0.567 & 0.750 \\ & PB & 0.771 & 0.451 & 0.749 & 0.610 & 0.170 & 0.677 & 0.642 & 0.784 \\ & Theory & 0.791 & 0.504 & 0.776 & 0.542 & 0.087 & 0.784 & 0.641 & 0.773 \\ & Combined & 0.804 & 0.515 & 0.777 & 0.661 & 0.155 & 0.715 & 0.687 & 0.810 \\
 \bottomrule
\end{tabular}
\begin{tablenotes}
    \small
    \item Note: Survey = survey model; PB = past-behavior model; Theory = theory-based model; SVM = sup-port vector machine; Acc = accuracy; TPR = true positive rate; FPR = false positive rate; NPV = negative prediction value; MMC = Matthews correlation coefficient.
    \end{tablenotes}
\end{table}

\begin{figure}[ht]
  \centering
  \includegraphics[width=0.95\linewidth]{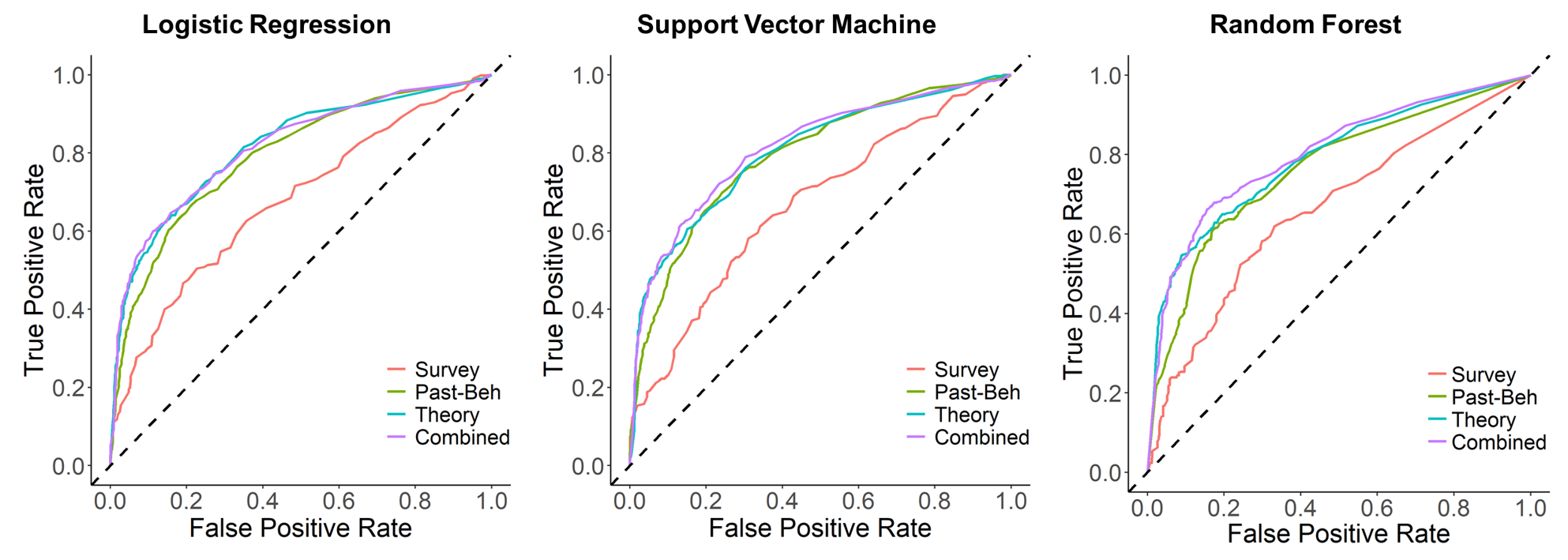}
  \setlength{\belowcaptionskip}{-5pt}
  \caption{Model comparison results of Study 2 based on ROC curves for different models and algorithms.}
  \label{fig:result2}
\end{figure}

\subsection{Performance across the two datasets}
\label{new}
Because all three algorithms gave similar results and logistic regression was the best overall, we only report the results of logistic regression models. Results of the models’ abilities for predicting unseen data from a different study are summarized in Figure 8 and Table 4 \footnote{When predicting Study 2's data, thresholds used for the logistic regression models were 0.57 (Survey), 0.56 (PB), 0.49 (Theory), and 0.62 (Combined). When predicting Study 1's data, thresholds used for the logistic regression models were 0.53 (Survey), 0.33 (PB), 0.3 (Theory), and 0.27 (Combined)}. For all models except the theory-based model, prediction performance when predicting data from a different study was close to the corresponding performance when predicting unseen subsets from the same study. However, the performance of theory-based model dropped quite a bit when predicting data from a different study (e.g., AUC decreased from 0.737 to 0.718 for Study 1, and from 0.815 to 0.753 for Study 2), suggesting that parameters in the theory-based equations (e.g., HDP and HGP) tuned using one data set were not necessarily the optimal for a different study. 

\begin{table}
\centering
\caption{Comparison of model performances in predicting new data.}
\label{tab:results}
\renewcommand{\arraystretch}{1.2}
\begin{tabular}{ c l c c c c c c c c } 
 \toprule
  & & AUC & MCC & Acc & TPR & FPR & Precision & F1-score & NPV \\
 \midrule
 \multirow{4}{*}{\rotatebox[origin=c]{90}{\parbox[c]{1.2cm}{\centering Predicting data set 2}}} & Survey & 0.680 & 0.282 & 0.655 & 0.601 & 0.313 & 0.529 & 0.563 & 0.746 \\ & PB & 0.793 & 0.448 & 0.740 & 0.673 & 0.221 & 0.641 & 0.667 & 0.803 \\ & Theory & 0.753 & 0.372 & 0.704 & 0.628 & 0.252 & 0.593 & 0.610 & 0.775 \\ & Combined & 0.795 & 0.477 & 0.763 & 0.598 & 0.141 & 0.713 & 0.650 & 0.785 \\ 
 \midrule
 \multirow{4}{*}{\rotatebox[origin=c]{90}{\parbox[c]{1.2cm}{\centering Predicting data set 1}}} & Survey & 0.677 & 0.262 & 0.603 & 0.383 & 0.149 & 0.742 & 0.505 & 0.551 \\ & PB & 0.733 & 0.375 & 0.688 & 0.684 & 0.307 & 0.714 & 0.698 & 0.661 \\ & Theory & 0.718 & 0.298 & 0.648 & 0.638 & 0.340 & 0.678 & 0.658 & 0.619 \\ & Combined & 0.749 & 0.352 & 0.678 & 0.750 & 0.403 & 0.676 & 0.711 & 0.680 \\
 \bottomrule
\end{tabular}
\begin{tablenotes}
    \small
    \item Note: Survey = survey model; PB = past-behavior model; Theory = theory-based model; Acc = accuracy; TPR = true positive rate; FPR = false positive rate; NPV = negative prediction value; MMC = Matthews correlation coefficient.
    \end{tablenotes}
\end{table}

\begin{figure}[ht]
  \centering
  \includegraphics[width=0.65\linewidth]{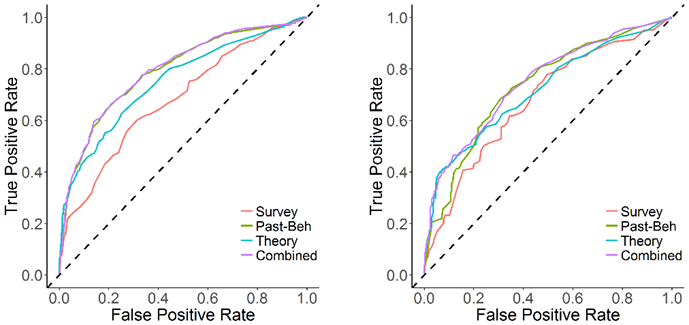}
  \setlength{\belowcaptionskip}{-5pt}
  \caption{Model comparison results in terms of predicting new data, based on ROC curves of differ-ent models (Left panel: predicting Study 2’s data using models trained on Study 1’s data; Right panel: predicting Study 1’s data using models trained on Study 2’s data).}
  \label{fig:new}
\end{figure}

\subsection{Parameter Estimation}
\label{parameter}
Lastly, for theoretical interests, we examined the optimal parameter values for the free parameters in the theory-based equations of habit strength and accessibility. For parameters governing the dynamics of habit strength, optimal ranges of parameter values could be found, and the results were similar regardless of the data set used (see Figure 9). To achieve best performance based on AUC, the optimal value for the habit decay parameter (\textit{HDP}) was in the range of 0.15 and 0.2, while the optimal value for the habit gain parameter (\textit{HGP}) was in the range of 0.1 and 0.2.

In contrast, for parameters that determine the dynamics of accessibility, there was no clear relationships between their values and model prediction performance (see Figure 10). If one examined the individual features in the theory-based models, the feature habit strength contributed to most of their predictive powers, while the feature accessibility did not contribute as much.

\begin{figure}[ht]
  \centering
  \includegraphics[width=0.7\linewidth]{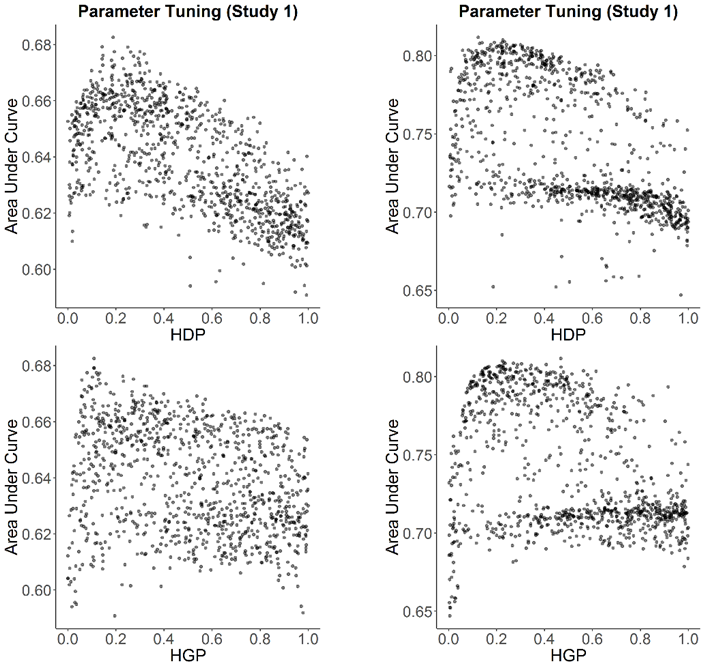}
  \setlength{\belowcaptionskip}{-5pt}
  \caption{Tuning results for parameter \textit{HDP} and \textit{HGP} in the computational model of habit strength, shown as the relationship between parameter values (x-axis) and model per-formance (area under curve, y-axis).}
  \label{fig:habit}
\end{figure}

\begin{figure}[ht]
  \centering
  \includegraphics[width=0.7\linewidth]{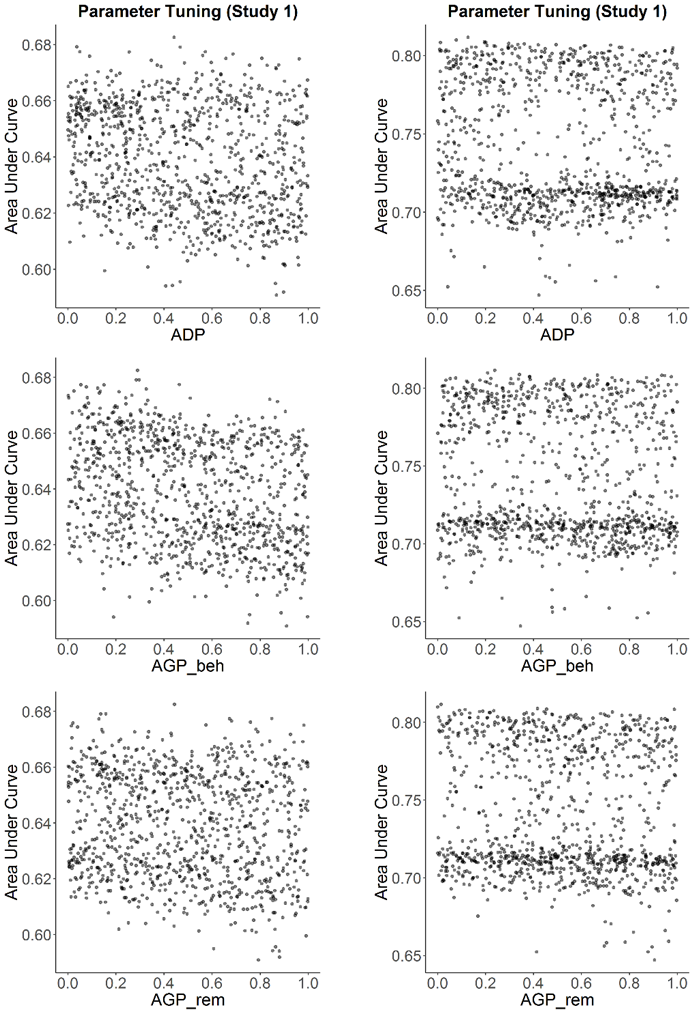}
  \setlength{\belowcaptionskip}{-5pt}
  \caption{Tuning results for parameter \textit{ADP}, \textit{$AGP_{beh}$}, and \textit{$AGP_{rem}$} in the computational model of memory accessibility, shown as the relationship between parameter values (x-axis) and model per-formance (area under curve, y-axis).}
  \label{fig:memory}
\end{figure}

\section{General Discussion}
\label{discussion}
Recently developed theory-based computational models allow BCSS to model users' habit learning in behavior change processes. In this paper, we reviewed the computational models of habit learning and evaluated the utility of one of the models in a use case of behavior prediction, based on data collected in two field intervention studies on toothbrushing behavior. Through a nested cross-validation procedure, theory-based models with computed habit strength and memory accessibility were compared with two baseline models in terms of how well they could predict brushing behavior on the next day. In both studies, the theory-based models performed better than the survey models that used self-reported behavioral determinants as features. In the second larger study, the theory-based models also performed slightly better than the models that are based simply on past behavior rate. However, as suggested by the cross-study prediction results, the small advantage of using the theory-based approach comes with a cost that for predicting behavior in a different context and/or with different users, the free parameters in the theory-based equations need to be tuned again. One cannot assume that the same rates of habit formation and decay generalize to all application situations. 

We initially thought that the computed cognitive variables would increase the predictive power of model based on other commonly used features, such as behavioral determinants (e.g., attitude) and contextual factors (e.g., emotional states, environmental cues). Therefore, we were surprised that combining the computed habit strength with either instrumental or affective attitude self-reported in the surveys did not perform better than the theory-based models alone (not reported in the result sections). Of course, the measurements of attitude were on the weekly level, so it is yet to know whether knowing more immediate contextual factors would further increase prediction accuracy of brushing behavior (e.g., sleepiness of the person in the evening, behavior of the partner, etc.). Without knowing this information, the current prediction accuracy of around 65 - 77\% might be the limit.

Although the equation of habit strength was motivated by theories (e.g., \cite{klein2011computational,miller2019habits}), the computed variable also represents a specific summary of past behavior. The similarity between the theory-based models and the past-behavior models was also reflected in the fact that they seemed to provide similar information, since adding these features together did not improve performance much further. Compared with past-behavior models that weight each behavior in the past equally, the equation of habit strength weights behavior at different time point in the past in a more sophisticated way. Given the habit decay parameter, the contributions of behaviors that are far in the past to current habit strength are discounted in an exponential way, given by the decay parameter to the power of n ($HDP^n$), where n denotes the number of time steps to the past. Behaviors in the later stage of habit formation also tend to have increasingly smaller immediate contributions to the current habit strength because the habit gain parameter is modulated by the term 1-$HS_t$. For the purpose of behavior prediction alone, it would be interesting to examine more closely the mathematical properties of the equation and to explore whether other ways of weighing past behaviors could result in better prediction performance.

Besides the interest in behavior prediction, the parameter estimation procedure used in our studies also has implications for the theoretical understanding of habit formation. The optimal values tuned for the habit gain parameter are very close to the corresponding values of 0.19 obtained through a statistical modeling of the temporal dynamics of self-reported habit strength or behavioral automaticity \cite{lally2010habits}. However, inconsistent with previous studies that suggested much smaller habit decay parameter \cite{tobias2009changing,lally2010habits}, its value was in the same range with the habit gain parameter. In general, these results speak to the theoretical meaningfulness of the computational model of habit strength used for prediction. In contrast, the parameters in the equation of accessibility did not seem to have optimal values, which casts doubts onto the validity of modeling memory accessibility in its current form.

While our current findings are limited in the context of two intervention trials, our theory-based approach can be easily implemented in real BCSS. As long as behaviors and other contextual variables are observed by sensors and parameter values are estimated from existing data, a digital system can update its representations of the user's habit strength after every relevant behavioral context, without asking the user to report it repeatedly. Real-time behavior prediction using computed habit strength provides the basis for delivering personalized and adaptive interventions. Instead of classifying brushing and non-brushing, the system can simply estimate the probability of brushing (non-brushing) and then use different thresholds for delivering different types of interventions. For example, if brushing probabilities stay very low for several days (e.g., 10\%), the system may decide to repeat an extensive education session about the importance of optimal oral health routine. Instead, if a user is predicted to brush the next morning with a probability of 0.6, a gentle reminder may be sent. Such adaptive interventions are important because even though the costs of delivering digital interventions are low, too frequent or inappropriate actions may disrupt or even irritate users \cite{mehrotra2016my}. Besides behavior prediction, a system may use the computed habit strength more directly. For example, tracking a user’ habit strength of a newly trained behavior may give the system a better idea about the progress of behavior change. Even when the target behavior is already performed consistently, a habit strength weaker than a certain threshold (e.g., 0.8) would suggest that the current intervention should be continued to reduce the risk of relapse. Future research should explore these different ways of implementing our theory-based approach in real-world  applications and extend our work to other behavioral domains beyond oral health. 

\bibliographystyle{unsrt} 
\bibliography{references} 
\end{document}